\pdfoutput=1

\documentclass[sigconf,nobalancelastpage]{acmart}

\usepackage{balance}

\def\BibTeX{{\rm B\kern-.05em{\sc i\kern-.025em b}\kern-.08emT\kern-.1667em\lower.7ex\hbox{E}\kern-.125emX}}

\copyrightyear{2019} 
\acmYear{2019} 
\setcopyright{acmcopyright}
\acmConference[KDD '19]{The 25th ACM SIGKDD Conference on Knowledge Discovery and Data Mining}{August 4--8, 2019}{Anchorage, AK, USA}
\acmPrice{15.00}
\acmDOI{10.1145/3292500.3330864}
\acmISBN{978-1-4503-6201-6/19/08}

\usepackage{algorithmic}

\usepackage{subfigure}

\usepackage{graphicx}

\usepackage{amsmath}
\usepackage{subfigure}
\usepackage{amsfonts,amssymb}

\usepackage{nicefrac}       

\usepackage{color}
\usepackage[linesnumbered,ruled,vlined]{algorithm2e}
\newtheorem{theorem}{Theorem}[section]
\newtheorem{lemma}[theorem]{Lemma}

\newtheorem{corollary}[theorem]{Corollary}

\usepackage{pgfplots}
\usepackage{pgfplotstable}
\pgfplotsset{width=7cm,compat=1.13}

\begin{document}

	\title{Off-policy Learning for Multiple Loggers}


\author{Li He}
\affiliation{\institution{JD.com}}
\email{heli@amss.ac.cn}

\author{Long Xia}
\affiliation{\institution{JD.com}}
\email{xialong@jd.com}

\author{Wei Zeng}
\affiliation{\institution{Institute of Computing Technology, CAS}}
\email{zengwei@software.ict.ac.cn}

\author{Zhi-Ming Ma}
\affiliation{\institution{Academy of Mathematics and Systems Science, CAS}}
\email{mazm@amt.ac.cn}

\author{Yihong Zhao}
\affiliation{\institution{JD.com}}
\email{ericzhao@jd.com}

\author{Dawei Yin}
\affiliation{\institution{JD.com}}
\email{yindawei@acm.org}

%
\renewcommand{\shortauthors}{Trovato and Tobin, et al.}

%

\begin{abstract}

It is well known that the historical logs are used for evaluating and learning policies in interactive systems, e.g. recommendation, search, and online advertising. Since direct online policy learning usually harms user experiences, 
it is more crucial to apply off-policy learning in real-world applications instead. Though there have been some existing works, most are focusing on learning with one single historical policy. However, in practice, usually a number of parallel experiments, e.g. multiple AB tests, are performed simultaneously. To make full use of such historical data, learning policies from multiple loggers becomes necessary. Motivated by this, in this paper, we investigate off-policy learning when the training data coming from multiple historical policies. Specifically, policies, e.g. neural networks, can be learned directly from multi-logger data, with counterfactual estimators. In order to understand the generalization ability of such estimator better, we conduct generalization error analysis for the empirical risk minimization problem. We then introduce the generalization error bound as the new risk function, which can be reduced to a constrained optimization problem. Finally, we give the corresponding learning algorithm for the new constrained problem, where we can appeal to the minimax problems to control the constraints. Extensive experiments on benchmark datasets demonstrate that the proposed methods achieve better performances than the state-of-the-arts.
\end{abstract}

%
%


%

%
\maketitle

\section{Introduction}
In many interactive systems, such as search engines, recommender systems, and ad-placement ~\cite{DBLP:journals/jmlr/BottouPCCCPRSS13, DBLP:journals/corr/AtheyW17}, large batches of logs are collected from the past periods for model improvement. 
Usually, the interactive process can be formulated as follows. Given an input~(or context)~from users, the system draws an output~(or action)~based on its current policy. Then we observe feedback of the chosen output for that input. The logged data are counterfactual since they only provide partial information. For example, in a movie recommendation system, we can only observe the feedback for the output chosen by the system (e.g. the recommended movie) but not for all the other movies that the system could have recommended potentially. Although the logs are biased, they are informative and can be exploited for many purposes. 

One application of logged data is \textit{off-policy evaluation}, which is to evaluate new given policies offline~\cite{Li:2011:UOE:3045725.3045727}. Since the online evaluation is prohibitively expensive and may harm the user experiences, leveraging such logged data could provide a useful alternative. Nevertheless, direct evaluation over the logged data which are collected from a historical policy, leads to a biased estimation. To address this issue, many estimators have been proposed~\cite{10.1093/biomet/70.1.41,doi:10.1198/106186008X320456,DBLP:conf/icml/DudikLL11,DBLP:conf/nips/SwaminathanJ15}.
Another important application of such logs is to learn policy with better performance, also known as \textit{off-policy learning} \cite{DBLP:journals/jmlr/BottouPCCCPRSS13,DBLP:conf/icml/SwaminathanJ15,DBLP:conf/icml/WuW18}. For example, the counterfactual estimator had been used for learning in advertisement applications~\cite{DBLP:journals/jmlr/BottouPCCCPRSS13}.

Though there have been some existing works, most are focusing on learning with one single historical policy. However, in practice, usually a number of parallel experiments, e.g. multiple AB tests, are performed simultaneously. This typically generates the logged data from many policies. To make full use of such historical data, learning policies from multiple loggers becomes an important problem. 
More recently, there have been few preliminary investigations for the case of multiple loggers. The work~\cite{DBLP:conf/kdd/AgarwalBSJ17} proposed three estimators for off-policy evaluation, which are named as naive, balanced, and weighted inverse propensity score~(IPS) estimators. However, it focuses on the evaluation problem but not on the learning. Following ~\cite{DBLP:conf/kdd/AgarwalBSJ17}, a weighted counterfactual risk minimization (WCRM) principle was proposed in \cite{Su2018}, which combined the weighted IPS estimator with counterfactual risk minimization principle. While they give an empirical sample variance regularization, it results in a more difficult optimization due to its dependency on the whole training data. 

In this paper, we 
investigate off-policy learning in the setting of multiple loggers. Specifically, we study the popular multi-logger counterfactual estimators, and apply them to the off-policy learning, where the popular model, e.g. neural networks based policies, could be learned directly. 
To better leverage such counterfactual estimators, we conduct generalization error analysis for the empirical risk minimization problem. Then, based on the analysis, we propose to optimize the generalization error bound 
instead to improve the generalization ability. 
The optimization of generalization error bound is further reduced to a constrained optimization problem, which is not uncommon and is kind of related with distributionally robust learning ~\cite{DBLP:conf/nips/NamkoongD17, DBLP:journals/mp/BertsimasGK18}. Finally, we propose the learning algorithm for the new constrained problem, appealing to the minimax problems for the constraints. We evaluate our new methods on three benchmark datasets. 
Empirically, our new methods perform better than the state-of-the-arts, and the results confirmed the theoretical analyses. 

To sum up, the main contributions of this paper are summarized as follows.

\begin{itemize}
	\item Theoretically, we conduct generalization error analyses for popular counterfactual estimators in multi-logger setting; 
	
	\item Based on the generalization error analyses, we use the generalization error bound as the new risk objectives and formulate them into constrained optimization problems;
	
	\item We provide corresponding learning algorithms for the new constrained optimization problems, appealing to
	the minimax problems to control the constraints;
	
	\item Empirically, we carry out experiments and analyses on three benchmark datasets. The results show that our new methods improve over the state-of-the-art methods.
\end{itemize}

The rest of the paper is organized as follows. We introduce the background of off-policy learning from multiple historical policies and review some related works in Section~\ref{sec:bg}.  We conduct the generalization error analysis for $\lambda$-weighted risk estimator in Section~\ref{sec:ge}. The constrained learning methods and their corresponding algorithms are proposed in Section~\ref{sec:lm}. Experiments are reported in Section~\ref{sec:exp} and a variant of estimator is provided in Section~\ref{sec:var}.

\section{Preliminaries}\label{sec:bg}
In this section, we describe the off-policy learning problem of multiple loggers and review some related works.

\subsection{Problem Setting}

We first recall how an interactive system works. Specifically, given an input~(or context)~ $x\in \mathcal{X}$, which is drawn from an unknown distribution $P(\mathcal{X})$, the system selects an output~(or action)~ $y\in \mathcal{Y}$ based on existing policy $h_0(\mathcal{Y}|x):\mathcal{X}\mapsto \mathcal{Y}$. 
We denote the probability assigned by $h_0(\mathcal{Y}|x)$ to $y$ as $h_0(y|x)$.
Then we observe feedback for the couple $(x,y)$ from an unknown function $\delta: \mathcal{X} \times \mathcal{Y} \rightarrow [0,L]$. 
The lower the value of $\delta(x,y)$, the higher the user satisfaction with this given output $y$ for the input $x$.

To perform the off-policy evaluation, we need to consider a specific risk objective. 
Usually, the risk of a new given policy $h(\mathcal{Y}|x)$ can be defined as
\[R(h) = \mathbb{E}_{x\sim P(\mathcal{X}), y\sim h(\mathcal{Y}|x)}\left[\delta(x,y) \right].\]
Due to the distribution mismatch between the policy $h(\mathcal{Y}|x)$ and the historical policy $h_0(\mathcal{Y}|x)$, we apply the importance sampling technique ~\cite{10.1093/biomet/70.1.41,DBLP:conf/nips/CortesMM10}. 
Therefore, the risk can be rewritten as
\[R(h) =\mathbb{E}_{x\sim P(\mathcal{X}), y\sim h_0(\mathcal{Y}|x)}\left[\frac{h(y|x)}{h_0(y|x)}\delta(x,y)\right].\] 
In addition, since the distribution $P(\mathcal{X})$ is unknown, we have to use the empirical estimator.
Assume we have a dataset from the historical policy $h_0(\mathcal{Y}|x)$, 
denoted as  
\[\mathcal{D} = \{(x_1, y_1, \delta_1,p_1),\dots, (x_{n}, y_{n}, \delta_{n},p_{n})\}\]
where $\delta_i\equiv \delta(x_i, y_i)$ and $p_i\equiv h_0(y_i|x_i)$, $i\in\{1,2,\dots,n\}$. 
We can use the following unbiased empirical estimator
\[\hat{R}(h)=\frac{1}{n}\sum_{i=1}^n  \frac{h(y_i|x_i)}{h_0(y_i|x_i)} \delta_i,\] 
for the expected loss $R(h)$.
This is the widely used inverse propensity score (IPS) approach ~\cite{10.1093/biomet/70.1.41}.

In this paper, we study off-policy learning in the multi-logger setting, i.e., learning a policy that has low risk by using the logs from multiple policies. This is practical and necessary as the policy gets updated frequently in most online systems. 
Denote the logs obtained from each logging policy $h_j$ with
\[\mathcal{D}^j = \left\{(x_1^j, y_1^j, \delta_1^j,p_1^j),\dots, (x_{n_j}^j, y_{n_j}^j, \delta_{n_j}^j,p_{n_j}^j)\right\},\]
where $x_i^j\sim P(\mathcal{X})$ and $y_i^j\sim h_j(\mathcal{Y}|x_i^j)$ for $j\in\{1,2,\dots,J\}, i \in\{1,2,\dots,n_j\}$. The feedback $\delta_i^j\equiv \delta(x_i^j,y_i^j)$ and the logging probability $p_i^j\equiv h_j(y_i^j|x_i^j)$. Therefore, we obtain a larger dataset $\mathcal{D}\equiv \cup_{j=1}^J \mathcal{D}^j$. Note that we can assume that all of the logging policies have the same input and output spaces. For simplicity, we denote $[J] = \{1, 2, \dots, J\}$ and $n=\sum_{j=1}^J n_j$.
Sometimes, we will use the abbreviations $h$ and $h_j$ for $h(y|x)$ and $h_j(y|x)$, respectively. 

Let us briefly review two recent related works below .

\subsubsection {Direct Learning Principle} 

In the case of multiple loggers, the work 
~\cite{DBLP:conf/kdd/AgarwalBSJ17} proposed an estimator for off-policy evaluation, which is called naive IPS estimator. 
Its definition is 
as follows.

\noindent {\bf Naive IPS Estimator}
\[\hat{R}_{naive}(h)=\frac{1}{n}\sum_{j=1}^J\sum_{i=1}^{n_j}\frac{h(y_i^j|x_i^j)}{h_j(y_i^j|x_i^j)}\delta_i^j,\]
This is an unbiased estimator when the logging policies have a full support for the target policy. 

However, as stated in~\cite{DBLP:conf/kdd/AgarwalBSJ17}, it suffers when they diverge to a degree where throwing away data lowers the estimator's variance.
As a result, they proposed $\lambda$-weighted IPS estimator, which is also unbiased but has a smaller variance than the naive IPS estimator. 

\noindent {\bf $\lambda$-Weighted IPS Estimator}

\[\hat{R}_{\lambda}(h)=\sum_{j=1}^J \lambda_j \sum_{i=1}^{n_j}\frac{h(y_i^j|x_i^j)}{h_j(y_i^j|x_i^j)}\delta_i^j,\]
where $\lambda_j\geq 0$ and $\sum_{j=1}^J\lambda_j n_j=1$. When taking $\lambda_j=\frac{1}{n}$, it reduces to the naive IPS estimator. If imposing $\lambda_j=\lambda_j^* \equiv \frac{1}{\sigma_{\delta}^2(h||h_j)\sum_{j=1}^J\frac{n_j}{\sigma_{\delta}^2(h||h_j)}}$, it becomes the weighted IPS estimator. The term $\sigma_{\delta}^2(h||h_j)$ is defined as the divergence from $h_j$ to $h$ in terms of the naive IPS estimator variance, which is estimated by the empirical divergence estimator in ~\cite{DBLP:conf/kdd/AgarwalBSJ17}. 

Once having an estimator, we can carry out off-policy learning by solving 
\begin{equation}\label{op:direct}
	\min_h	\hat{R}_{\lambda}(h). \nonumber
\end{equation}
While it is simple and natural, direct learning from the counterfactual estimators and choosing the empirical optimal minimizer still have several pitfalls ~\cite{DBLP:conf/icml/SwaminathanJ15}, 
such as it may have vastly different variances for two different loggers. 
An improved learning method for multiple loggers is needed.
In previous works~\cite{DBLP:conf/icml/SwaminathanJ15,Su2018}, the learned policies are usually formulated as stochastic softmax rules. 
In this paper, we adopt neural network to express the learned policy.

\subsubsection{WCRM Principle}

Following the above work, a weighted counterfactual risk minimization~(WCRM)~principle was proposed in ~\cite{Su2018}. The WCRM objective is
\begin{align}
	&
	\arg\min \sum_{j=1}^J\lambda_j^*\sum_{i=1}^{n_j}u_i^j(h)+\lambda\sqrt{\frac{\hat{Var}\left(\lambda_j^*n_j u_i^j(h)\right)}{n}},\nonumber\\
	&\hat{Var}\left(\lambda_j^*n_j u_i^j(h)\right)=\frac{1}{n-1}\sum_{j=1}^J\sum_{i=1}^{n_j}\left(\lambda_j^*n_ju_i^j-\frac{1}{n}\sum_{j=1}^J\sum_{i=1}^{n_j}\lambda_j^*n_ju_i^j\right)^2, \nonumber
\end{align}where 
$u_i^j(h)=\frac{h(y_i^j|x_i^j)}{h_j(y_i^j|x_i^j)}\delta_i^j$. 
The divergence $\sigma_{\delta}^2(h||h_j)$ is estimated by a self-normalized divergence estimator. The variance term in WCRM principle depends on the whole dataset, which results in a more difficult stochastic optimization. 

%
%

\subsection{Related Work}

Our paper is related to off-policy evaluation, which has many applications in practice, such as recommendation, search engine, and learning to rank~\cite{DBLP:conf/wsdm/LiCLW11,DBLP:journals/corr/LiCKG14, DBLP:conf/sigir/JoachimsS16, DBLP:conf/nips/SwaminathanKADL17, DBLP:conf/ijcai/JoachimsSS18, DBLP:conf/wsdm/WangGBMN18}.  
The counterfactual estimator can date back to the inverse propensity score (IPS) estimator ~\cite{HorvitzA1952generalization, 10.1093/biomet/70.1.41}.
It uses importance weighting technique, which solves the 
mismatch between the training distribution and the test distribution. In ~\cite{DBLP:conf/nips/CortesMM10}, the authors firstly give the theoretical learning bound analysis for importance weighting. 
However, the IPS estimator may have large or unbounded variance.
As a result, many new estimators have arisen, such as truncated importance sampling ~\cite{doi:10.1198/106186008X320456}, doubly robust estimator ~\cite{DBLP:conf/icml/DudikLL11}, and self-normalized estimator ~\cite{DBLP:conf/nips/SwaminathanJ15}. 
The doubly robust estimator was first developed for regression ~\cite{cassel1976some}, then it had been brought to contextual bandits.  
The self-normalized estimator was developed to avoid propensity overfitting problem. Both doubly robust and self-normalized estimators fall into the method of control variate ~\cite{mcbook2013owen}. 
Recently, researchers proposed many counterfactual estimators with smaller mean square error ~\cite{su2018continuous, DBLP:journals/corr/abs-1809-03084}.

The off-policy evaluation can be regarded as counterfactual reasoning for analyzing the causal effect of a new treatment from previous data ~\cite{DBLP:journals/jmlr/BottouPCCCPRSS13,DBLP:conf/icml/ShalitJS17}. 
It can also be viewed as a special case of off-policy evaluation in reinforcement learning ~\cite{DBLP:journals/tnn/SuttonB98, DBLP:conf/icml/PrecupSS00, DBLP:conf/icml/JiangL16, DBLP:conf/nips/MunosSHB16,DBLP:conf/icml/ThomasB16, DBLP:conf/icml/FarajtabarCG18}, which also has been applied in real applications~\cite{zou2019reinforcement, DBLP:journals/corr/abs-1801-00209,zhao2019, DBLP:conf/recsys/ZhaoXZDYT18, DBLP:conf/kdd/ZhaoZDXTY18,  DBLP:journals/corr/abs-1902-03987,zou2019RLkdd}.
Exploiting logs is important in multi-armed bandit and its variants, such as contextual bandit ~\cite{DBLP:conf/nips/StrehlLLK10, DBLP:journals/jmlr/ShivaswamyJ12}. 

The counterfactual estimator is often the first step for learning problem. 
For off-policy learning, 
there are also some related works~\cite{DBLP:conf/kdd/BeygelzimerL09, 	DBLP:conf/nips/StrehlLLK10, DBLP:journals/jmlr/BottouPCCCPRSS13,DBLP:conf/icml/SwaminathanJ15, DBLP:conf/nips/SwaminathanJ15, DBLP:conf/icml/WuW18}. 
For example, the work ~\cite{DBLP:journals/jmlr/BottouPCCCPRSS13} used the counterfactual estimator for learning in advertisement application. 
In ~\cite{DBLP:conf/icml/SwaminathanJ15}, the authors developed the counterfactual risk minimization (CRM) principle for batch learning from bandit feedback (BLBF). 
The key idea lies in controlling the differences in variance between different target policies. 
They derived an algorithm, policy optimizer for exponential models (POEM) , for learning stochastic linear rules for structured output prediction. 
The work~\cite{DBLP:conf/nips/SwaminathanJ15} proposed the Norm-POEM algorithm by combining the self-normalized estimator with POEM algorithm.
The work that is the most related to ours is ~\cite{DBLP:conf/kdd/AgarwalBSJ17}, in their work, they pointed out the sub-optimality of the naive IPS estimator and proposed two alternative estimators: balanced IPS estimator and weighted IPS estimator.  Recently, in~\cite{Su2018}, the authors combined the weighted IPS estimator with counterfactual risk minimization principle for learning from multiple loggers.

\section{Generalization Error Analysis}\label{sec:ge}
Now that we have an empirical risk minimization (ERM) problem, we can study its generalization error bound, which is frequently-used in supervised learning. In this section, we give the generalization error analyses for $\lambda$-weighted IPS estimator, while the analysis for naive IPS estimator can be obtained by letting $\lambda_j=\frac{1}{n}$.

Before conducting the generalization error analysis, we first borrow one lemma from ~\cite{DBLP:conf/nips/CortesMM10} as follows.

\begin{lemma}\label{lem:bound}
	(~\cite{DBLP:conf/nips/CortesMM10}) For a random variable $z$, let $Q$ and $Q_0$ be two probability measures, assume $q$ and $q_0$ are two probability density functions of $Q$ and $Q_0$, respectively. Let $l$ be a loss function bounded in $[0,1]$. Let $w = \frac{q}{q_0}$,  then the following 
	results hold:
	\begin{align*}
		&\mathbb{E}_{z\sim Q_0}[w(z)] = 1,\ \  \mathbb{E}_{z\sim Q_0}[w^2(z)] = d_2(q||q_0),\\  
		&\mathbb{E}_{z\sim Q_0}\left[w^2(z)l^2(z)\right] \leq d_2(q||q_0),
	\end{align*}
	where $d_2(q||q_0)\equiv 2^{D_2(q||q_0)}$ and $D_2(q||q_0)$ is R\'enyi divergence~\cite{renyi1961}. 
\end{lemma}

Based on lemma \ref{lem:bound}, we can obtain an upper bound for the second moment of the importance weighted loss, i.e.,
\begin{align*}
	\mathbb{E}_{x\sim P(\mathcal{X}), y\sim h_0(\mathcal{Y}|x)}
	\left[\left(\frac{h(y|x)}{h_0(y|x)}\delta(x,y)\right)^2\right]
	\leq L^2 d_2(h(y|x)||h_0(y|x); P(x)),
\end{align*}
where  $d_2(h(y|x)||h_0(y|x);P(x))
\triangleq d_2(P(x)h(y|x)||P(x)h_0(y|x))= \int_{\mathcal{X}, \mathcal{Y}} \frac{h^2(y|x)}{h_0(y|x)}P(x) \mathrm{d}x\mathrm{d}y.$
This can be easily derived by some substitutions, i.e., letting $z=(x,y), q(z)=P(x)h(y|x), q_0(z)=P(x)h_0(y|x)$, and $l(z)=\delta(x,y)\in[0, L]$, so here we omitted the proof~\citep{DBLP:conf/icml/WuW18}.

Now we are ready to give our main theorem and the sketch of the proof as follows.   

\begin{theorem}\label{th:weighted}
	Let $R(h)$ be the risk of a new policy $h$ on the loss function $\delta$, and $\hat{R}_{\lambda}(h)$ be the $\lambda$-weighted empirical risk. 
	Assume the divergence is bounded by $M_{j}$, i.e., $d_2(h||h_j) \leq d_{\infty}(h||h_j) = M_{j}$~\footnote{The divergence $d_{\infty}(h||h_j) \equiv \sup_{y}\frac{h(y|x)}{h_j(y|x)}$.}, $j\in[J]$ and denote $M_{\lambda}\equiv max_j\{\lambda_j M_j\}$. Then, for any $\eta>0$, with probability at least $1-\eta$, the following bound holds:
	{\small
		\begin{equation*}
			R(h) \leq  \hat{R}_{\lambda}(h)+\frac{2L M_{\lambda}\log\frac{1}{\eta}}{3} +L\sqrt{2\sum_{j=1}^J n_j\lambda_j^2  d_2(h||h_j;P(x))\log \frac{1}{\eta}}.
		\end{equation*}
	}
\end{theorem}

{\it Proof.} By the definition of $\lambda_j$, we have 
{\small
	\begin{align*}
		R(h)-\hat{R}_{\lambda}(h)=\sum_{j=1}^J \sum_{i=1}^{n_j}\lambda_j\left[R(h)-\frac{h(y_i^j|x_i^j)}{h_j(y_i^j|x_i^j)}\delta(x_i^j,y_i^j)\right],
	\end{align*}
}Denote $Z_i^j=R(h)-\frac{h(y_i^j|x_i^j)}{h_j(y_i^j|x_i^j)}\delta(x_i^j,y_i^j)$ and $Z=R(h)-\frac{h(y|x)}{h_j(y|x)}\delta(x,y)$. 
We can obtain that $\mathbb{E}_{x\sim P(\mathcal{X}), y\sim h_j(\mathcal{Y}|x)}Z=0$ and 
$\left|Z \right|
\leq M_j L$.

In addition, by applying lemma \ref{lem:bound}, we have 
{\small 
	$$\mathbb{E}_{x\sim P(\mathcal{X}), y\sim h_j(\mathcal{Y}|x)} \left[\left(\frac{h(y|x)}{h_j(y|x)}\delta(x,y)\right)^2\right]
	\leq L^2 d_2(h(y|x)||h_j(y|x);P(x))
	.$$
}Thus, we have the following bound for the second moment of $Z$,
\begin{equation*}
	\mathbb{E}_{x\sim P(\mathcal{X}), y\sim h_j(\mathcal{Y}|x)} Z^2
	\leq L^2 d_2(h(y|x)||h_j(y|x);P(x)).
\end{equation*}

Applying Bernstein's inequality ~\cite{bennett1962probability}, we have
{\small
	\begin{align*}
		&\mathbb{P}\left(\sum_{j=1}^J\sum_{i=1}^{n_j}\lambda_j Z_i^j>\epsilon\right)\\
		&\leq \exp\left(-\frac{\frac{1}{2}\epsilon^2}{\sum_{j=1}^J\sum_{i=1}^{n_j}
			\mathbb{E}_{x\sim P(\mathcal{X}), y\sim h_j(\mathcal{Y}|x)} (\lambda_j Z_i^j)^2+\frac{1}{3}L M_{\lambda}\epsilon}\right)\\
		&\leq \exp\left(-\frac{\frac{1}{2}\epsilon^2}{\sum_{j=1}^J\sum_{i=1}^{n_j}\lambda_j^2 L^2 d_2(h(y|x)||h_j(y|x);P(x))+\frac{1}{3}L M_{\lambda}\epsilon}\right),
	\end{align*}
}where $M_{\lambda}=max_j\{\lambda_j M_j\}$. 

Let the right hand be equal to $\eta$, we can obtain an quadratic function of $\epsilon$. With some calculations and using $\sqrt{a+b}\leq\sqrt{a}+\sqrt{b}$, we obtain 
that
{\small $$\epsilon\leq \frac{2L M_{\lambda}\log \frac{1}{\eta}}{3}+L\sqrt{2\sum_{j=1}^J n_j\lambda_j^2 d_2(h(y|x)||h_{j}(y|x);P(x))\log\frac{1}{\eta}}.$$}
Therefore, the following inequality
{\small
	\begin{align*}
		&R(h)\leq \hat{R}_{\lambda}(h)\\
		&+ \frac{2L M_{\lambda}\log \frac{1}{\eta}}{3}+L\sqrt{2\sum_{j=1}^J n_j\lambda_j^2 d_2(h(y|x)||h_{j}(y|x);P(x))\log\frac{1}{\eta}}
	\end{align*}
}holds with probability at least $1-\eta$. \ \ \ \  $\Box$

When taking $\lambda_j$ as $\lambda_j^*$, we obtain the generalization error bound of weighted IPS estimator. When letting $\lambda_j=\frac{1}{n}, \forall j$, $\lambda$-weighted IPS estimator reduces to the naive IPS estimator. Therefore, we give the following corollary.

\begin{corollary}\label{th:naive}
	Let $R(h)$ be the risk of a new policy $h$ on the loss function $\delta$, and $\hat{R}_{naive}(h)$ be the naive empirical risk. 
	Assume that the divergence is bounded by $M_{j}$, i.e., $d_2(h||h_j) \leq d_{\infty}(h||h_j) = M_{j}, j\in[J]$ and denote $M_{naive}\equiv max_j\{M_j\}$. 
	Then, for any $\eta>0$, with probability at least $1-\eta$, the following bound holds:
	{\small
		\begin{align*}
			R(h) \leq &\hat{R}_{naive}(h)\\
			&+\frac{2L M_{naive}\log\frac{1}{\eta}}{3n} +L\sqrt{\frac{2\sum_{j=1}^J n_j d_2(h||h_j;P(x))\log \frac{1}{\eta}}{n^2}}.
		\end{align*}
	}
\end{corollary}

In the next section, we show how to apply the generalization error analyses to the off-policy learning.

\section{Constrained Learning Methods}\label{sec:lm}
In this section, we introduce our new constrained learning methods and study how to apply them in practice. 
Based on the results in Section~\ref{sec:ge}, we propose to use the generalization error bound as the new regularized objectives. Specifically, we have
{\small
	\begin{equation}\label{eq:weighted}
		\min_{h}\hat{R}_{weighted}(h)+\beta\sqrt{\sum_{j=1}^J n_j(\lambda_j^*)^2 d_2(h||h_j;P(x))},
\end{equation}}
{\small
	\begin{equation}\label{eq:naive}
		\min_{h}\hat{R}_{naive}(h)+\beta\sqrt{\frac{\sum_{j=1}^J n_jd_2(h||h_j;P(x))}{n^2}},
	\end{equation}
}for the weighted IPS estimator and the naive IPS estimator, respectively. The parameter $\beta=\sqrt{2L^2\log\frac{1}{\eta}}$
is to trade-off the bias and the regularization, which is challenging to be set empirically and solving the optimization problem. 
Thus, inspired by the work~\cite{DBLP:conf/icml/WuW18}, we study a constrained optimization problem instead.
For eq.(\ref{eq:weighted}) and eq.(\ref{eq:naive}), we consider the following general constrained problem 
{\small
	\begin{align}\label{op:weighted}
		& \min_{h} \sum_{j=1}^J\lambda_j\sum_{i=1}^{n_j}\frac{h(y_i^j|x_i^j)}{h_j(y_i^j|x_i^j)}\delta_i^j \\ 
		& s.t.  \sum_{j=1}^J n_j\lambda_j^2d_2(h||h_j;P(x))\leq\frac{\rho}{n^2}, \nonumber
\end{align}}where $\rho$ is a pre-defined constant. 
Similarly, it corresponds to the formulation of the naive IPS estimator with $\lambda_j=\frac{1}{n}, j\in[J]$.

In the next sections, we first review the derivations of variational divergence minimization. Then we give our algorithms for the proposed constrained formulations.

\subsection{About the Constraints}
For eq.(\ref{op:weighted}), we have to analyze the constraint, where the term $d_2(h||h_j;P(x))$ is defined as 
\[d_2(h||h_j;P(x))= \int_{\mathcal{X}, \mathcal{Y}} \frac{h^2(y|x)}{h_j(y|x)}P(x) \mathrm{d}x\mathrm{d}y.\]
With some derivations, we can obtain 
\begin{equation}
	d_2(h||h_j;P(x)) = D_f(h||h_j;P(x))+1,\nonumber
\end{equation}where $D_f(h||h_j;P(x))=\int_{\mathcal{X}}D_f(h||h_j)P(x)\mathrm{d}x$ and $f(t)=t^2-1$.
Here the term $D_f(h||h_j)$ is the $f$-divergence ~\cite{DBLP:journals/ftcit/CsiszarS04}.  Hence, if we are able to control the part of $D_f(h||h_j;P(x))$ well, 
we can obtain the upper bound of $d_2(h||h_j;P(x))$ immediately. Thus, we omit the constant $1$ without loss of generality. 

By following the techniques in ~\cite{DBLP:journals/tit/NguyenWJ10, DBLP:conf/nips/NowozinCT16}, we obtain the lower bound of $D_f(h||h_j;P(x))$,
\begin{equation}\label{eq:bound}
	D_f(h||h_j;P(x)) \geq 
	sup_{T\in\mathcal{T}}\left\{\mathbb{E}_{x,y\sim h}T(x,y)-\mathbb{E}_{x,y\sim h_j}f^*(T(x,y))\right\},
\end{equation}where $f^*$ is a convex conjugate function of $f$~\cite{fenchel2012}~\footnote{The convex conjugate function is defined as $f^*(t)=\sup_{u\in dom_f}\{ut-f(u)\}$.}, and $\mathcal{T}$ is an arbitrary class of functions $T: \mathcal{X} \rightarrow \mathbb{R}$. 
For the inequality, under mild conditions on $f$ ~\cite{DBLP:journals/tit/NguyenWJ10}, the bound is tight for $T^*(x)=f'(\frac{h}{h_j})$, where $f'$ is the first derivative of $f$. 
On the other hand, we can choose $T$ as the family of neural networks to obtain tight bound which benefits from the universal approximation theorem~\cite{DBLP:journals/nn/HornikSW89}. 

Therefore, in order to control the constraint, we need to solve a minimax problem. More specifically, we first maximize the lower bound for the establishment of eq.(\ref{eq:bound}), then we minimize the $f$-divergence by choosing the optimal $h$. We represent $T$ function as a discriminative network parameterized with $w$ and express the policy $h(\mathcal{Y}|x)$ as a generator network with parameter $\theta$. Define $F(\theta, w^j)$ as
\begin{equation}\label{eq:f}
	F(\theta, w^j) = \mathbb{E}_{x,y\sim h_{\theta}}T_{w^j}(x,y)-\mathbb{E}_{x,y\sim h_j}f^*(T_{w^j}(x,y)). 
\end{equation}
To optimize eq.(\ref{eq:f}) on a finite training dataset, we can use mini-batch version to approximate the expectation. To approximate $\mathbb{E}_{x,y\sim h_j}[\cdot ]$, we sample $B$ instances without replacement from the training set. To approximate $\mathbb{E}_{x,y\sim h_{\theta}}[\cdot ]$, we can sample $B$ instances from the current generative policy $h_{\theta}$.

\subsection{Learning Algorithms}\label{sec:alg}

For constrained naive and weighted learning methods, we propose their algorithms in Algorithms~ \ref{alg:3.1}--\ref{alg:3.2}. 
\begin{algorithm}[t]
	\renewcommand{\algorithmicrequire}{\textbf{Input:}}
	\renewcommand{\algorithmicensure}{\textbf{Output:}}
	\caption{Constrained Naive and Weighted Learning Algorithm}
	\label{alg:3.1}
	\begin{algorithmic}[1]
		\REQUIRE Dataset $\mathcal{D}^j,j\in[J]$, learning rate $\eta_1$, threshold $\rho$, max iteration $I$, max epoches $MAX$
		\ENSURE Optimized generator $h_{\theta_*}(y|x)$ that is an approximate minimizer of $R(h)$\\
		initialization
		\STATE \textbf{Repeat}
		\STATE Sample a mini-batch of $B$ real samples $(x_i^j, y_i^j)$ from $\mathcal{D}^j$
		\STATE Calculate $\hat{R}^{mini}=n\frac{1}{JB} \sum_{j=1}^J \lambda_j\sum_{i=1}^B \frac{h_{\theta_t}(y_i^j|x_i^j)}{h_j(y_i^j|x_i^j)}\delta_i^j$ and the gradient $g=\partial_{\theta}\hat{R}^{mini}$
		\STATE Update $\theta_{t+1}=\theta_t-\eta_1 g$
		\STATE Call Algorithm \ref{alg:3.2} to minimize $\sum_{j=1}^J n_j \lambda_j^2\hat{D}_f(h||h_j;P(x))$ with threshold $\frac{\rho}{n^2}$ and max iteration $I$
		\STATE \textbf{Until} epoch $> MAX$
	\end{algorithmic}  
\end{algorithm}
Correspondingly, for $\lambda_j=\frac{1}{n},\forall j\in[J]$, it becomes the constrained naive learning algorithm, and if letting $\lambda_j=\lambda_j^*$, it turns to the learning algorithm of weighted IPS estimator.

In Algorithm~\ref{alg:3.2}, we leverage the Gumbel-softmax estimator in step 3. For structured output problem with discrete values, the gradients of samples obtained from the distribution $h(\mathcal{Y}|x)$ cannot backpropagate to all other parameters. The works~\cite{Jang2016Gumbel} and ~\cite{DBLP:journals/corr/MaddisonMT16} developed a continuous relaxation of discrete random variables in stochastic computational graphs, which can generate approximated and differentiable samples. The main ideas are as follows. They first use Gumbel-Max trick to represent a multinomial distribution, then it can be approximated by Gumbel-softmax distribution. Mathematically, given a categorical distribution with class probabilities $\pi_1, \pi_2, \dots, \pi_k$, the Gumbel-softmax estimator generates an approximate one-hot sample $y$ with
\begin{equation}
	y_i=\frac{\exp\left(\nicefrac{\left(\log(\pi_i\right)+ g_i)}{\tau}\right)}{\sum_{j=1}^k \exp\left(\nicefrac{\left(\log(\pi_j\right)+ g_j)}{\tau}\right)},  i=1,\dots,k,
\end{equation}where $\tau$ is the temperature and $g_1,\dots,g_k$ are i.i.d samples drawn from $Gumbel(0,1)$ distribution. 
The term $\hat{D}_f(h||h_j;P(x))$ denotes {\small \[\frac{1}{B}\sum_{(x_i^j,\hat{y}_i^j)\sim h_{\theta_t}}T_{w_t^j}(x,y)-\frac{1}{B}\sum_{(x_i^j,y_i^j)\sim h_{j}}f^*(T_{w_t^j}(x,y)),\]
}i.e., the mini-batch version of $F(\theta_t,w_t^j)$.

\begin{algorithm}[t]
	\renewcommand{\algorithmicrequire}{\textbf{Input:}}
	\renewcommand{\algorithmicensure}{\textbf{Output:}}
	\caption{Variational Minimization of the Constraint} 
	\label{alg:3.2}
	\begin{algorithmic}[1]
		\REQUIRE Dataset $\mathcal{D}^j,j\in[J]$, threshold $D$, an initial generator $h_{\theta_0} (y|x)$, discriminator function $T_{w_{0}^j}(x,y),j\in[J]$, learning rates $\eta_h, \eta_T$, max iteration $I$
		\ENSURE Optimized generator $h_{\theta_*}(y|x)$ which satisfies the constraint\\
		initialization
		\STATE \textbf{Repeat}
		\STATE Sample a mini-batch of $B$  real samples $(x_i^j, y_i^j)$ from $\mathcal{D}^j$ for each $j\in[J]$
		\STATE Sample a mini-batch of $B$ input $x$ from $\mathcal{D}^j$ for each $j\in[J]$, and construct fake samples $(x_i^j,\hat{y}_i^j)$  by sampling from $h_{\theta^t}(y|x)$ with Gumbel-softmax sampling 
		\STATE Update $\theta_{t+1}=\theta_t-\eta_h \partial_{\theta}\left(\sum_{j=1}^J n_j \lambda_j^2F(\theta_t, w_t^j)\right)$
		
		\STATE Update\\
		$w_{t+1}^j 
		= w_t^j+\eta_T \partial_{w^j}\left(n_j \lambda_j^2 F(\theta_t, w_t^j)\right)$, $j\in[J]$	
		
		\STATE \textbf{Until} $\sum_{j=1}^J n_j \lambda_j^2\hat{D}_f(h||h_j;P(x))\le D$ or iteration $>I$
	\end{algorithmic}  
\end{algorithm}
\section{Experiments}\label{sec:exp}
In this section, we empirically evaluate the proposed algorithms, i.e., naive and weighted constrained algorithms on three benchmark datasets.

\subsection{Experimental Settings}
We first introduce the datasets and  the experimental methodology.

\subsubsection{Datasets and Methodology}
In our experiments, we choose multi-label classification task due to the access of a rich feature space and an easily scalable label space. Three multi-label datasets are collected from the LibSVM repository ~\cite{DBLP:journals/tist/ChangL11} for the following experiments. Each dataset consists of feature $x\in \mathbb{R}^{p_F}$ and its corresponding supervised label $y^*\in\{0,1\}^{q_L}$. The datasets have different feature dimension $p_F$, label dimension $q_L$, and sample number $n$. Statistics on the datasets are given in Table \ref{tb:1}. For the dataset TMC, since it has sparse features with high dimension, we reduce the feature dimension to 1000 via truncated singular value decomposition~(latent semantic analysis).

\begin{table}  
	\begin{center}  
		\caption{Statistics on Scene, Yeast, and TMC.}  
		\label{tb:1}
		\begin{tabular}{c c c c c}  
			\hline Dataset & $p_F$ ($\#$feature) & $q_L$ ($\#$label) & $n_{train}$ & $n_{test}$ \\  
			\hline   
			Scene & 294 & 6  &1211   & 1196\\ 
			Yeast & 103 & 14 &1500   &917  \\  
			TMC & 30438 & 22 & 21519 & 7077  \\  
			\hline  
		\end{tabular}  
	\end{center}  
\end{table}

To control the experiments more effectively, we derive bandit data from these three full-information datasets. We follow the supervised $\mapsto$ bandit conversion method in ~\cite{DBLP:conf/icml/AgarwalHKLLS14}.
For supervised data $\mathcal{D}^*=\{(x_i,y_i^*)\}_{i=1}^n$, we first train the conditional random fields~(CRF)~method ~\cite{DBLP:conf/icml/LaffertyMP01} on a part of $\mathcal{D}^*$ to obtain logging policies. For the simplest setting, CRF actually performs logistic regression for each label independently. Following ~\cite{DBLP:conf/icml/SwaminathanJ15}, we consider using the stochastic softmax rules 
{\small
	\begin{equation}
		h^w(y|x)=\frac{\exp(w^T\phi(x,y))}{\mathbb{Z}(x)}\nonumber
\end{equation}}as the hypothesis space.
The $\phi(x,y)$ is the joint feature map of input $x$ and output $y$, and $\mathbb{Z}(x)=\sum_{y'\in\mathcal{Y}}\exp(w^T\phi(x,y'))$ is the partition function.
We also use a stochastic multiplier $\alpha$ in the map of $w\mapsto \alpha w$ to control the "stochasticity" of the logging policies, where larger $\alpha$ will induce a more deterministic variant of $h^w$.  

For simplicity and ease of interpretability, we consider two logging policies in the following experiments.
We first train a CRF on 20\% of data, then scale the parameter $w \mapsto \alpha w$ with $\alpha=0.05$ to obtain logger $h_1$. The second logger $h_2$ is trained on the same data with stochastic multiplier to be $2$. 
To create bandit feedback datasets $\mathcal{D}\equiv \mathcal{D}^1 \cup \mathcal{D}^2$, we take $4$ passes through $\mathcal{D}^*$ and sample labels by simulating $y_i^j\sim h_j(x_i)$.
We use the Hamming loss as the loss function $\delta(x_i^j,y_i^j)$, which is the number of incorrectly assigned labels between the sampled label $y_i^j$ and the supervised label $y_i^*$.
For the test dataset, we report the expected loss per test instance 
\[EXP = \frac{1}{n_{test}}\sum_{i=1}^{n_{test}}\mathbb{E}_{y\sim h(\mathcal{Y}|x_i)}[\delta(y_i^*, y)]\] for the learned policy $h(\mathcal{Y}|x)$.

\subsubsection{Baselines and Implementations}
We compare our constrained naive and weighted algorithms, denoted by \textbf{Naive-Reg} and \textbf{Weighted-Reg}, with the following baselines:
\begin{itemize}
	\item \textbf{WCRM}: We compare with the weighted counterfactual risk minimization (WCRM) principle ~\cite{Su2018}. They used the clipped version of estimator and applied the limited memory-BFGS (L-BFGS) algorithm ~\cite{DBLP:journals/mp/LewisO13} from scikit-learn ~\cite{DBLP:journals/jmlr/PedregosaVGMTGBPWDVPCBPD11} for optimization. We conduct it by following their experimental descriptions. 
	\item \textbf{Naive/Weighted}: We utilize neural networks to present the policies for direct naive and weighted learning principles in Section~\ref{sec:bg}.
\end{itemize}
For references, we also report the results from a supervised CRF~(denoted as \textbf{CRF}), and the expected Hamming losses of $h_1$~(denoted as \textbf{Logger 1}) and $h_2$~(denoted as \textbf{Logger 2}). All CRF, Logger 1 and Logger 2 are actually built in supervised learning, where \textbf{CRF} is trained on the whole training dataset and the stochastic multiplier $\alpha$ is set as 1.
We follow the implementations in ~\cite{DBLP:conf/icml/SwaminathanJ15} and the details can be found in their paper.

We keep aside 25\% of the bandit dataset $\mathcal{D}$ as the validation set. The 
$EXP$ is chosen according to the performance of the validation loss. 
For loggers and WCRM principle, we follow the experimental setup in ~\cite{DBLP:conf/icml/SwaminathanJ15,Su2018}. 
Direct learning principle and our algorithms are implemented with TensorFlow~\footnote{https://www.tensorflow.org/} in the experiments. We use Adam ~\cite{Kingma2014adam} to train the neural networks. The learning rate of the re-weighted loss is set as $0.0001$. For the regularization part, we set $0.0001$ for the learned policy network. The learning rates for the discriminative networks are not fixed, one need to adjust it and we usually choose $0.00025$. We set the batch size as $500$ for Yeast and Scene datasets, and $4096$ for TMC dataset. Some detailed configurations are put in the appendix. 
In addition, we leverage the Gumbel-softmax estimator for differential sampling, which was developed for variational methods ~\cite{Jang2016Gumbel,DBLP:journals/corr/MaddisonMT16}. For 
the weight $\lambda_j$, we apply the self-normalized divergence estimator in ~\cite{Su2018}, i.e., 
{\small
	\begin{align*}
		&\tilde{\sigma}_{\delta}^2(h||h_j)=\frac{1}{n_j-1}\sum_{i=1}^{n_j}\left(\frac{u_i^j(h)}{S^j(h)}-\bar{u}(h)\right)^2,\\
		& S^j(h) = \frac{1}{n_j}\sum_{i=1}^{n_j}\frac{h(y_i^j|x_i^j)}{h_j(y_i^j|x_i^j)},\ \  \bar{u}(h) = \frac{1}{n}\sum_{j=1}^J\sum_{i=1}^{n_j}u_i^j(h).
	\end{align*}
}

\subsection{Experimental Results}

\begin{table}  
	\begin{center}  
		\caption{The comparisons of the expected Hamming loss on three datasets.}  
		\label{tb:2}\setlength{\tabcolsep}{4mm}{
			\begin{tabular}{c c c c} 
				\hline   Method       &  Scene   & Yeast      & TMC  \\  
				\hline  Logger 1     & 2.866    & 6.898      &  10.322   \\  
				Logger 2     & 0.960    & 4.306      &  2.014   \\  
				\hline  WCRM         & 1.088    & 3.908      & 4.232    \\
				\hline  Naive         & 1.056    & 4.001      & 4.452    \\
				\textbf{Naive-Reg}    & 1.037    & 3.551      & \textbf{2.415}    \\
				\hline  Weighted      & 1.011    & 3.756      & 3.041    \\
				\textbf{Weighted-Reg} & \textbf{0.994}      & \textbf{3.263}     & 2.748   \\
				\hline CRF           & 0.942  & 4.133     & 1.612   \\ 
				\hline  
		\end{tabular} } 
	\end{center}  
\end{table}

The expected Hamming loss $EXP$ on three datasets are reported in Table~\ref{tb:2}.
Lower loss is better.
From the results we can see that, Naive-Reg beats the baseline of WCRM and Naive, while Weighted-Reg beats the baseline of WCRM and Weighted. Not surprised, Weighted-Reg achieves better performance than Naive-Reg in Scene and Yeast data set. 
The results indicate that our constrained learning method is effective due to the improvement of the generalization ability. 
Although Naive-Reg/Weighted-Reg can not surpass the better logger~(i.e., Logger 2, trained in supervised learning), both of them perform much better than the baselines.
We also notice that in some cases~(e.g. on Yeast dataset), Weighted-Reg even gets competitive results, against supervised CRF method which is trained on the whole training dataset.


\subsection{Experiments on Varying Replay Count}

In this section, we aim to explore how the constrained naive and weighted algorithms work with varying replay count. 

In the previous section, we both take $4$ passes through $\mathcal{D}^*$ and sample labels for the two loggers. The stochastic multipliers are still set as 0.05 and 2 for logger $h_1$ and $h_2$, respectively. Keeping $4$ passes for logger $h_2$, we vary the number of times that we replayed the training set~(replay count)~to collect labels from logger $h_1$ over $\{2^0, 2^1, 2^2, 2^3, 2^4\}$. The purpose of varying $h_1$ replay count rather than $h_2$, is to see the performance change of two proposed algorithms Naive-Reg and Weighted-Reg under different proportions of stochastic data (recall that $h_1$ has more explorations). 

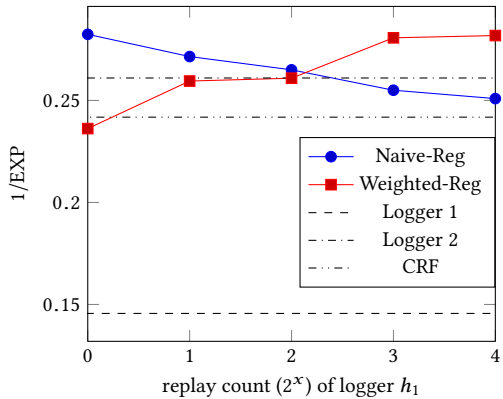
\begin{figure}
	\centering
	\begin{tikzpicture}\small
	\begin{axis}[
	xmin=0, xmax=4,
	xlabel={replay count ($2^x$) of logger $h_1$},
	ylabel={$1/\mathrm{EXP}$},
	legend style={at={(1,0.61)}}]
	\addplot
	coordinates{
		(0,0.2824) (1,0.2715) (2,0.2650) (3,0.2550) (4,0.2509)
	};
	\addlegendentry{Naive-Reg}
	\addplot
	coordinates
	{
		(0,0.2362) (1,0.2595) (2,0.2609) (3,0.2807) (4,0.2818)
	};
	\addlegendentry{Weighted-Reg}
	\addplot[mark=none, black, dashed]{0.1456}; 
	\addlegendentry{Logger 1}
	\addplot[mark=none, black, dashdotted]{0.2610}; 
	\addlegendentry{Logger 2}
	\addplot[mark=none, black, dashdotdotted]{0.2418};
	\addlegendentry{CRF}
	\end{axis}
	\end{tikzpicture}
	\caption{Generalization performance of Naive-Reg and Weighted-Reg as varying $h_1$ replay count on the Yeast dataset.}
	\label{fig:replay}
\end{figure}

We report the results of Logger 1 and Logger 2, CRF, Naive-Reg and Weighted-Reg on Yeast dataset in Figure~\ref{fig:replay}. The horizontal axis denotes the replay count number and the vertical axis is the inverse of the expected Hamming loss. The blue curve denotes the constrained naive method and the red curve denotes the constrained weighted method.


As shown in the figure, the performance of constrained naive algorithm gets worse with the increasing replay count. 
As mentioned above, smaller multiplier would lead to more stochastic logging policy. 
Logger 1 is more stochatic and adding more data of it will dilute those information from Logger 2.
Intuitively, this would deteriorate the performance of the learned policy. Whereas the cases are different for the constrained weighted algorithm, it performs better along with the increasing replay count. Since the constrained weighted learning method assigns diﬀerent weights for loggers, it can take advantage of the growing training data size and get rid of the effects from stochastic data. 

\subsection{Experiments on Varying Temperature}

As mentioned in sec.\ref{sec:alg}, we leverage the Gumbel-softmax trick for differential sampling. There is a temperature parameter $\tau$ in the Gumbel-softmax estimator, in this section, we study whether our learning methods are robust to this parameter.

We conduct the following analyses on two bandit datasets generated from Yeast. To eliminate the effects of parameters, the parameters for constrained naive learning method are set to be same with direct learning, except those parameters for the regularization. Furthermore, we keep the same parameters for all of the constrained naive learning methods except for the varying temperature $\tau$.
Specifically, naive method use one hidden layer with 10 hidden nodes for the learned policy network. The learning rate is set as 0.0001 and the batch size is set as 500. For the regularization, we also adopt one hidden layer but with 59 hidden nodes. The learning rates are set to be 0.0001 and 0.00025 for the step 4 and step 5 in algorithm~\ref{alg:3.2}, respectively.

For constrained weighted learning, we hold the same parameter set with that of direct weighted learning. All of the constrained weighted learning methods use identical structure for the regularization except for the varying temperature $\tau$. Specifically, weighted method has two hidden layers with 7 nodes in each hidden layer. The learning rate is also set as 0.0001 and the batch size is 500. For the regularization, we choose to use one hidden layer with 30 nodes,  and also set 0.0001 and 0.00025 for step 4 and step 5 in algorithm~\ref{alg:3.2}.

\begin{figure}
	\begin{tikzpicture}\small
	\begin{axis}[
	xmin=0, xmax=5,
	ymin=0.22, ymax=0.3,
	xlabel={temperature ($\tau$)},
	ylabel={$1/\mathrm{EXP}$},
	legend pos=north west
	]
	\addplot[mark=none, blue]{0.2499};
	\addlegendentry{Naive}
	\addplot[mark=*,blue]
	coordinates{
		
		(0.5,0.2375) (1,0.2518) (1.5,0.2508) (2,0.2667) (2.5,0.2601) (3,0.2583) (3.5,0.2683) (4,0.2816) (4.5,0.2469) (5,0.2752)
	};
	\addlegendentry{Naive-Reg}
	
	\addplot[mark=none, blue, dashed]{0.2559};
	\addlegendentry{WCRM}
	\end{axis}
	\end{tikzpicture}
	\caption{Generalization performance of Naive-Reg as varying temperature parameter $\tau$ on the Yeast dataset.}
	\label{fig:naive-tem}
\end{figure}
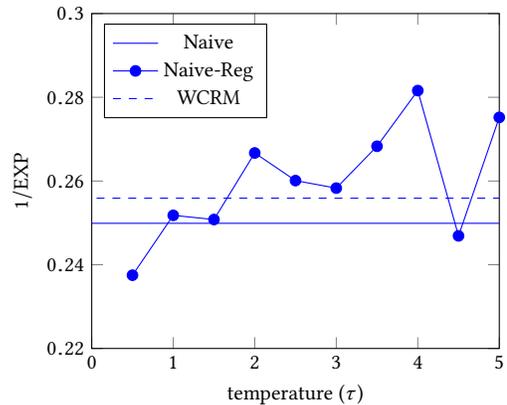

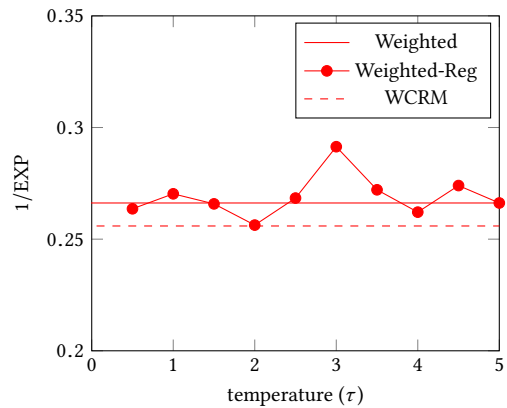
\begin{figure}
	\begin{tikzpicture}\small
	\begin{axis}[
	xmin=0, xmax=5,
	ymin=0.2, ymax=0.35,
	xlabel={temperature ($\tau$)},
	ylabel={$1/\mathrm{EXP}$},
	]
	\addplot[mark=none, red]{0.2662};
	\addlegendentry{Weighted}
	\addplot[mark=*,red]
	coordinates
	{
		(0.5,0.2636) (1,0.2703) (1.5,0.2658) (2,0.2563) (2.5, 0.2684) (3,0.2914) (3.5,0.2721) (4,0.2621) (4.5,0.2740) (5,0.2662)
	};
	\addlegendentry{Weighted-Reg}
	\addplot[mark=none, red, dashed]{0.2559};
	\addlegendentry{WCRM}
	\end{axis}
	\end{tikzpicture}
	\caption{Generalization performance of Weighted-Reg as varying temperature parameter $\tau$ on the Yeast dataset.}
	\label{fig:weighted-tem}
\end{figure}

We vary the temperature $\tau$ over $\{0.5, 1, 1.5, 2, 2.5, 3, 3.5, 4, 4.5, 5\}$ and report the experimental results in Figures \ref{fig:naive-tem}-\ref{fig:weighted-tem}. The figures include WCRM, the direct learning version, and the constrained version. The vertical axis still denotes the inverse of the expected Hamming loss. As shown in the figures, we can see that the constrained versions are usually better than the direct learning versions. This shows that our new constrained learning methods are stable with the Gumbel-softmax temperature parameter.

\section{Other Variants of Estimators}\label{sec:var}

Besides the $\lambda$-weighted IPS estimator, in this section, we introduce the balanced IPS estimator~\cite{DBLP:conf/kdd/AgarwalBSJ17}, provide its generalization error analysis and the constrained learning algorithm correspondingly.

\noindent {\bf Balanced IPS Estimator}
\[\hat{R}_{bal}(h)=\frac{1}{n}\sum_{j=1}^J\sum_{i=1}^{n_j}\frac{h(y_i^j|x_i^j)}{h_{avg}(y_i^j|x_i^j)}\delta_i^j,\]
where $h_{avg}(y|x) \equiv \frac{\sum_{j=1}^J n_j h_j(y|x)}{n}$. 
This is also an unbiased estimator with a smaller variance than naive IPS estimator.

\begin{theorem}\label{th:bal}
	Let $R(h)$ be the risk of a new policy $h$ on the loss function $\delta$, and $\hat{R}_{bal}(h)$ be the balanced empirical risk. 
	Assume that the divergence is bounded by $M_{avg}$, i.e., $d_2(h||h_{avg}) \leq d_{\infty}(h||h_{avg}) = M_{avg}$. 
	Then, for any $\eta>0$, with probability at least $1-\eta$, the following bound holds:
	{\small
		\begin{equation*}
			R(h) \leq \hat{R}_{bal}(h)+\frac{2LM_{avg}\log\frac{1}{\eta}}{3n} +L\sqrt{\frac{2d_2(h||h_{avg};P(x))\log \frac{1}{\eta}}{n}}.
		\end{equation*}
	}
\end{theorem}
If the last term is reformulated as 
$L\sqrt{\frac{2\sum_{j=1}^J n_j d_2(h||h_{avg};P(x))\log \frac{1}{\eta}}{n^2}}$, it becomes similar to that of the naive IPS estimator. 

We propose to use the following regularized objective, i.e.,
{\small
	\begin{equation}\label{eq:bal}
		\min_{h} \hat{R}_{bal}(h)+\beta\sqrt{\frac{d_2(h||h_{avg};P(x))}{n}}.
	\end{equation}
}Similarly, we minimize the following constrained optimization problem instead, i.e.,
{\small
	\begin{align}\label{op:bal}
		&\min_h \frac{1}{n}\sum_{j=1}^J\sum_{i=1}^{n_j}\frac{h(y_i^j|x_i^j)}{h_{avg}(y_i^j|x_i^j)}\delta_i^j\\   \nonumber 
		&s.t.   d_2(h||h_{avg};P(x))\leq\frac{\rho}{n}.
	\end{align}
}

We can give the corresponding algorithm for the constrained balanced learning method. 
Compared with Algorithm~\ref{alg:3.1}, we should replace step 3, i.e., the computation of minibatch gradient, with $\hat{R}^{mini}_{bal}=\frac{1}{JB}\sum_{j=1}^J\sum_{b=1}^B \frac{h_{\theta_t}(y_i^j|x_i^j)}{h_{avg}(y_i^j|x_i^j)}\delta(x_i^j, y_i^j)$. 
The step 2 in Algorithm~\ref{alg:3.2} should be modified by constructing samples from the policy $h_{avg}$. Since the quantities of $h_j(y_i^j|x_i^j)$ are assumed to be available for all possible $(x,y)$ pairs, we are able to calculate $h_{avg}(y_i^j|x_i^j)$ and sample from it. For the constrained balanced learning algorithm, we use $\hat{D}_f(h||h_{avg};P(x))$ to denote the mini-batch version of $F(\theta_t,w_t^{avg})$, i.e., 
{\small
	\begin{align*}
		&\hat{D}_f(h||h_{avg};P(x)) = \\
		&\frac{1}{JB}\sum_{(x_i^j,\hat{y}_i^j)\sim h_{\theta_t}}T_{w_t^{avg}}(x,y)-\frac{1}{JB}\sum_{(x_i^j,y_i^{avg})\sim h_{avg}}f^*(T_{w_t^{avg}}(x,y)).
	\end{align*}	
}Please refer to the appendix for the complete algorithm.

\section{Conclusion}\label{sec:con}
Preforming off-policy learning is becoming more important 
than online policy learning in real-world applications. Most of previous works are focused on the off-policy learning with one single historical policy.
In this paper, we studied the off-policy learning from multiple historical policies, which is important and realistic. 
The learned policy and the discriminative networks for learning are adopted as deep neural networks.
The generalization error analysis for the empirical risk minimization problem is provided. Based on the analysis, we proposed to use the generalization error bound as the new risk function, which can be alternatively transformed into a constrained optimization problem. Learning algorithm for the optimization problem is designed, through a minimax setting, to solve the constraint of the optimization problem. In experiments, we test the new methods on three benchmark datasets. Compared with direct learning principle and the WCRM principle, the performances of proposed algorithms outperform the state-of-the-art baselines. In the future, we will try to find other measures to control the differences between the logging policies and the learned policy.

\section*{Acknowledgements}
Zhi-Ming Ma was partially supported by National Center for Mathematics and Interdisciplinary Sciences of Chinese Academy of Sciences.

	%
\bibliographystyle{ACM-Reference-Format}
\balance
\bibliography{sample-base}

\newpage
%
\nobalance
\appendix
\section*{Appendix}
\section{Proofs of Theorem \ref{th:bal}}
\begin{theorem}
	Let $R(h)$ be the risk of a new policy $h$ on the loss function $\delta$, and $\hat{R}_{bal}(h)$ be the balanced empirical risk. 
	Assume that the divergence is bounded by $M_{avg}$, i.e., $d_2(h||h_{avg}) \leq d_{\infty}(h||h_{avg}) = M_{avg}$. 
	Then, for any $\eta>0$, with probability at least $1-\eta$, the following bound holds:
	{\small
		\begin{equation*}
			R(h) \leq \hat{R}_{bal}(h)+\frac{2LM_{avg}\log\frac{1}{\eta}}{3n} +L\sqrt{\frac{2d_2(h||h_{avg};P(x))\log \frac{1}{\eta}}{n}}.
		\end{equation*}
	}
\end{theorem}

{\it Proof.} By definition, we have 
{\small
	\begin{equation*}
		R(h)-\hat{R}_{bal}(h)
		=\frac{1}{n}\sum_{j=1}^J \sum_{i=1}^{n_j}\left[R(h)-\frac{h(y_i^j|x_i^j)}{h_{avg}(y_i^j|x_i^j)}\delta(x_i^j,y_i^j)\right].
	\end{equation*}
}Denote $X_i^j=R(h)-\frac{h(y_i^j|x_i^j)}{h_{avg}(y_i^j|x_i^j)}\delta(x_i^j,y_i^j)$ and 
$X=R(h)-\frac{h(y|x)}{h_{avg}(y|x)}\delta(x,y)$. 
Taking expectation, we have 
$\mathbb{E}_{x\sim P(\mathcal{X}), y\sim h_{avg}(\mathcal{Y}|x)} X=0$.
We can also derive that 
$$\left|X\right|
\leq \left|\frac{h(y|x)}{h_{avg}(y|x)}\delta(x,y)\right|
\leq M_{avg} L.$$
If $\frac{h(y|x)}{h_{avg}(y|x)}\geq M_{avg}$,
then $d_2(h||h_{avg}) \equiv \int_y \frac{h(y|x)}{h_{avg}(y|x)}h(y|x)dy
\geq \int_y M_{avg} h(y|x)dy=M_{avg}$. This contradicts with the assumption.

In addition, by applying lemma \ref{lem:bound}, we have 
{\small 
	\[\mathbb{E}_{x\sim P(\mathcal{X}), y\sim h_{avg}(\mathcal{Y}|x)} \left[\left(\frac{h(y|x)}{h_{avg}(y|x)}\delta(x,y)\right)^2\right]
	\leq L^2 d_2(h(y|x)||h_{avg}(y|x);P(x))
	.\]
}Thus, we have the following bound for the second moment of $X$,
\begin{equation*}
	\mathbb{E}_{x\sim P(\mathcal{X}), y\sim h_{avg}(\mathcal{Y}|x)} X^2
	\leq L^2 d_2(h(y|x)||h_{avg}(y|x);P(x)).
\end{equation*}

Applying Bernstein's inequality ~\cite{bennett1962probability}, we have
{\small
	\begin{align*}
		&\mathbb{P}\left(\frac{1}{n}\sum_{j=1}^J\sum_{i=1}^{n_j}X_i^j>\epsilon\right)\\
		&\leq \exp\left(-\frac{\frac{1}{2}n^2\epsilon^2}{\sum_{j=1}^J\sum_{i=1}^{n_j}\mathbb{E}_{x\sim P(\mathcal{X}), y\sim h_{avg}(\mathcal{Y}|x)} (X_i^j)^2+\frac{1}{3}L M_{avg}n\epsilon}\right)\\
		&\leq \exp\left(-\frac{\frac{1}{2}n^2\epsilon^2}{\sum_{j=1}^J n_j L^2 d_2(h(y|x)||h_{avg}(y|x);P(x))+\frac{1}{3}L M_{avg}n\epsilon}\right).
	\end{align*}
}Let the right hand be equal to $\eta$, then we can obtain that
{\small $$\log\frac{1}{\eta}=\frac{\frac{1}{2}n^2\epsilon^2}{\sum_{j=1}^J n_j L^2 d_2(h(y|x)||h_{avg}(y|x);P(x))+\frac{1}{3}L M_{avg}n\epsilon}.$$
}This is an quadratic function of $\epsilon$ and we solve that
{\small 
	$$\epsilon\leq \frac{2LM_{avg}\log \frac{1}{\eta}}{3n}+L\sqrt{\frac{2\sum_{j=1}^J n_j d_2(h(y|x)||h_{avg}(y|x);P(x))\log\frac{1}{\eta}}{n^2}}.$$}

Therefore, the following inequality
{\small
	\begin{align*}
		R(h)
		&\leq \hat{R}_{bal}(h)+ \\
		&\frac{2LM_{avg}\log \frac{1}{\eta}}{3n}+L\sqrt{\frac{2\sum_{j=1}^J n_jd_2(h(y|x)||h_{avg}(y|x);P(x))\log\frac{1}{\eta}}{n^2}}
	\end{align*}
}
holds with probability at least $1-\eta$.  \ \ \ \ $\Box$

\section{Constrained Balanced Learning Algorithm}

In this section, we will give the corresponding algorithm for the balanced IPS estimator.

\begin{algorithm}
	\renewcommand{\algorithmicrequire}{\textbf{Input:}}
	\renewcommand{\algorithmicensure}{\textbf{Output:}}
	\caption{Constrained Balanced Learning Algorithm}
	\label{alg:2.1}
	\begin{algorithmic}[1]
		\REQUIRE Dataset $\mathcal{D}^j,j\in[J]$ , threshold $D$, an initial generator $h_{\theta_0} (y|x)$, discriminator function $T^{avg}_{w_{0}}(x,y)$, max iteration $I$
		\ENSURE Optimized generator $h_{\theta_*}(y|x)$ that is an approximate minimizer of $R(w)$
		
		\STATE \textbf{Repeat}
		\STATE Sample each mini-batch of $B$ real samples $(x^j_i, y^j_i)$ from $\mathcal{D}^j,j\in[J]$
		
		\STATE Calculate $\hat{R}^{mini}_{bal}=\frac{1}{JB}\sum_{j=1}^J\sum_{i=1}^B \frac{h_{\theta_t}(y_i^j|x_i^j)}{h_{avg}(y_i^j|x_i^j)}\delta(x_i^j, y_i^j)$ and the gradient $g=\partial_{\theta}\hat{R}_{bal}^{mini}$
		
		\STATE Update $\theta_{t+1}=\theta_t-\eta_2 g$
		
		\STATE Call Algorithm \ref{alg:2.2} to minimize the divergence $\hat{D}_f(h||h_{avg};P(x))$ with threshold $\frac{\rho}{n}$, and max iteration $I$
		\STATE \textbf{Until} epoch $> MAX$
	\end{algorithmic}  
\end{algorithm}

\begin{algorithm}
	\renewcommand{\algorithmicrequire}{\textbf{Input:}}
	\renewcommand{\algorithmicensure}{\textbf{Output:}}
	\caption{Variational Minimization of the Constraint} 
	\label{alg:2.2}
	\begin{algorithmic}[1]
		\REQUIRE Dataset $\mathcal{D}^j,j\in[J]$, threshold $D$, an initial generator $h_{\theta_0} (y|x)$, discriminator function $T^{avg}_{w_{0} }(x,y)$, learning rates $\eta_h, \eta_T$, max iteration $I$
		
		\ENSURE Optimized generator $h_{\theta_*}(y|x)$ that has minimum divergence to $h_{avg}$
		
		\STATE \textbf{Repeat}
		
		\STATE Sample a mini-batch of $B$ real samples $(x^j_i, y^j_i)$ from $\mathcal{D}^j$ for each $j\in[J]$, and construct $JB$ samples $(x_i^j, y_i^{avg})$ by sampling from $h_{avg}(y|x)$
		
		\STATE Sample a mini-batch $B$ of input $x_i^j$ from $\mathcal{D}^j$ for each $j\in[J]$, and construct fake samples $(x_i^j,\hat{y}_i^j)$ by sampling from $h_{\theta_t}(y|x)$ with Gumbel-softmax sampling
		
		
		\STATE Update $\theta_{t+1}=\theta_t-\eta_h \partial_{\theta} F(\theta_t, w^{avg}_t)$
		
		\STATE Update $w^{avg}_{t+1}=w_t^{avg}+\eta_T \partial_{w^{avg}} F(\theta_t, w^{avg}_t)$

		\STATE \textbf{Until} $\hat{D}_f(h||h_{avg};P(x))\le D$ or iteration $>I$
	\end{algorithmic}  
\end{algorithm}

\section {Network Configurations} 
For the discriminative network $T$ and the generative network $h$, we use structures like below:

Discrimitive NN:  \textit{Linear$\to$  BatchNorm $\to$ ReLU $\to$  Linear $\to$  BatchNorm $\to$  ReLU $\to$ Linear} 

Generative NN:  \textit{Linear$\to$  BatchNorm $\to$  ReLU $\to$  Linear $\to$  BatchNorm $\to$  ReLU $\to$  Linear $\to$  Sigmoid}

\end{document}